%% file: paper-491.tex
\documentclass[runningheads]{llncs}
\usepackage{graphicx}

\usepackage{soul} 
\usepackage{times} 
\usepackage{epsfig} 

\usepackage{amsmath}
\usepackage{amssymb}
\usepackage{booktabs}

\usepackage{subcaption}
\usepackage{multirow}
\usepackage{float}
\usepackage{enumitem}
\usepackage{array}

\usepackage{pifont}
\usepackage{sidecap} 
\usepackage[breaklinks,colorlinks]{hyperref}

\newcommand{\xmark}{\ding{55}}%

\begin{document}
\title{Transformer based Multitask Learning for Image Captioning and Object Detection}
%
%
\author{Debolena Basak \thanks{corresponding author}
\and P.K. Srijith \and Maunendra Sankar Desarkar
}
\institute{Indian Institute of Technology Hyderabad, India\\
    \email{ai20resch11003@iith.ac.in, \{srijith, maunendra\}@cse.iith.ac.in }}
\maketitle  
\begin{abstract}
In several real-world scenarios like autonomous navigation and mobility, to obtain a better visual understanding of the surroundings, image captioning and object detection play a crucial role. This work introduces a novel multitask learning framework that combines image captioning and object detection into a joint model. We propose \textbf{TICOD}, \textbf{T}ransformer-based \textbf{I}mage \textbf{C}aptioning and \textbf{O}bject \textbf{D}etection model for jointly training both tasks by combining the losses obtained from image captioning and object detection networks. By leveraging joint training, the model benefits from the complementary information shared between the two tasks, leading to improved performance for image captioning. Our approach utilizes a transformer-based architecture that enables end-to-end network integration for image captioning and object detection and performs both tasks jointly. We evaluate the effectiveness of our approach through comprehensive experiments on the MS-COCO dataset. Our model outperforms the baselines from image captioning literature by achieving a $3.65\%$ improvement in BERTScore.

\end{abstract}
\section{Introduction}
\label{sec:intro}
\vspace{-6mm}
\begin{figure}[!]
    \centering
    \includegraphics[width=0.6\textwidth]{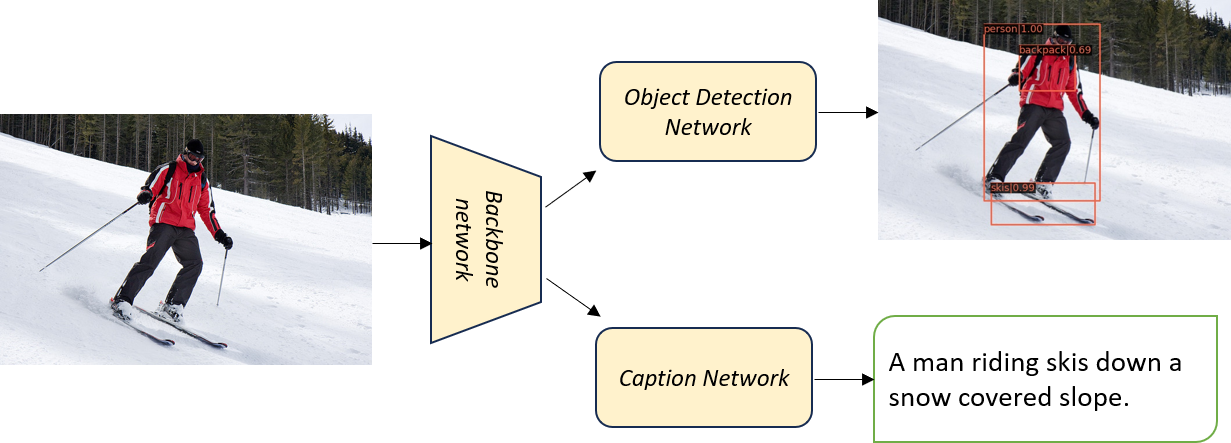}
    \caption{\small \textbf{A high-level framework of our proposed method}. Our model TICOD has three major components, which we call as --  (a) the \textit{backbone network}, (b) the \textit{object detection network}, and (c) the \textit{caption network.}}
    \label{fig: high-level-framework}
    \vspace{-6mm}
\end{figure}
In autonomous navigation and human mobility assistance systems, image captioning and object detection play a vital role in obtaining a better visual understanding of the surroundings. 
Applications such as human mobility assistance systems would require the detection of objects including their positions and description of the environment in natural language to help in the mobility of visually impaired people.

Most image captioning models follow the encoder-decoder framework along with the visual attention mechanism. The encoder processes input images and encodes them into fixed-length feature vectors, and the decoder utilizes these encoded image features to generate word-by-word descriptions \cite{ShowAndTell,ShowAttendAndTell,Anderson_Bottom-up,Pan_X-linear}. 
Most mainstream image captioning models~\cite{Anderson_Bottom-up,M2_transformer} use a two-step training method, which firstly extracts image regional features using a pre-trained object detector like Faster R-CNN~\cite{Faster-RCNN}, then feeds these feature vectors into an encoder-decoder framework for image captioning. However, this approach has a few inherent shortcomings: 1) the object detection model in the first step is trained on specific visual datasets like the Visual Genome, and the visual representation is not optimized towards the image captioning dataset used (commonly MS-COCO). This may lead to an error propagation problem if the object detector fails to recognize certain important visual information~\cite{Luo_AAAI}, 2) the time-intensive nature of extracting region features causes state-of-the-art models to rely on cached visual features (usually pre-computed) for training and evaluation, imposing constraints on model designs and resulting in run-time inference inefficiencies during prediction \cite{E2E-VLP,PureT,Jiang_2020_CVPR}. This introduces a two-stage process where a pre-trained object detection model and an image captioning model are used in sequence to extract features and predict the captions. The learning typically updates the image captioning model. This impedes the end-to-end training from image pixels to descriptions in image captioning models, limiting their applicability in real-world scenarios \cite{PureT,Jiang_2020_CVPR}.

Inspired by the NLP domain, the transformer architecture has shown its potential in computer vision (CV) tasks~\cite{ViT,Swin_transformer} and multimodal tasks~\cite{CLIP}. Considering the drawbacks of pre-trained CNN and object detector in the encoder and advantages of vision transformers, we integrate the task of image captioning as a single-stage approach, which can also perform object detection parallelly, as shown in Fig.~\ref{fig: high-level-framework}.
The key idea is \textit{multitask learning across object detection and image captioning} that enables the model to develop a better representation learning capability. This shared representation learning enables the model to leverage the knowledge gained from each task to effectively align the backbone representations, enhancing the overall learning capacity. The model parameters are learned by optimizing a  joint loss that combines the losses from both tasks. A key advantage of this approach is that if we want to generate a caption, we can simply enable the captioning network, turning off the object detection network, which helps us to get output without introducing any additional latency. On the other hand, if we need more detailed information,
we can enable the detection network to provide us with objects' details along with the captions simultaneously. 
Our model's generated captions and detected objects can be utilized to generate synthetic data using LLMs like GPT-4. For instance, a recent work \cite{LLaVA} has used COCO captions and detection annotations to feed into \textit{text-only} GPT-4 to generate \textit{multimodal instruction-following data}, which includes \textit{conversation}, \textit{detailed description} and \textit{complex reasoning} data. This synthetic data has been used for training the multimodal chatbot LLaVA~\cite{LLaVA}.

We use Swin Transformer \cite{Swin_transformer} as the backbone network for extracting image features. We use GPT2~\cite{GPT2} to decode the image features extracted from the Swin transformer and generate the captions of the corresponding image. For the object detection part, Cascade R-CNN~\cite{cascade_rcnn} framework is used with the same Swin backbone. We use the MS-COCO dataset \cite{COCO_objects,COCO_caps} for bounding box information and category labels to train object detection, and the caption annotations corresponding to the same image are used for image captioning together as joint training. 

To summarize, our key contributions are --
(a) We developed a \textbf{T}ransformer-based \textbf{I}mage \textbf{C}aptioning and \textbf{O}bject \textbf{D}etection  (\textbf{TICOD}) model capable of simultaneously performing image captioning and object detection, (b) TICOD uses a joint loss function that combines the losses from object detection and caption networks while maintaining a trade-off between them, (c) We demonstrate that our multitask method outperforms the task-specific models on image captioning by $3.65\%$ in BERTScore~\cite{bert-score} and produces comparable performance in object detection.
\vspace{-4mm}
\section{Related Work}
\label{sec_related_work}
\vspace{-2mm}
\noindent \textbf{Image Captioning:} 
Most of the existing image captioning works can be broadly categorized into CNN-RNN based models \cite{ShowAndTell,ShowAttendAndTell,Anderson_Bottom-up} and CNN-Transformer based models \cite{Pan_X-linear,M2_transformer,Luo_AAAI}. 
$\mathcal{M}^2$ Transformer~\cite{M2_transformer} used a mesh-like connection between each of the encoder and the decoder blocks to extract features from all levels of the encoder. It used Faster R-CNN \cite{Faster-RCNN} pre-trained on the Visual Genome dataset. The disadvantages of such an approach have been already discussed.
After Vision Transformer (ViT) \cite{ViT} and its variants \cite{Deit,Swin_transformer,CLIP} became popular in CV tasks, people began to explore it for image captioning as well. 
ClipCap \cite{clipcap} performs image captioning using a CLIP \cite{CLIP} encoder and GPT2 \cite{GPT2} decoder. 
Oscar \cite{oscar} and VinVL \cite{VinVL} use BERT \cite{BERT_devlin} but provide additional object tags for supervision, which limits their practical applicability in real-world scenarios. These approaches \cite{oscar,VinVL} are constrained to datasets that provide access to object detectors or annotations.
\textit{PureT} \cite{PureT} used Swin Transformer \cite{Swin_transformer} as a backbone network to extract image features for image captioning. 
They keep the Swin backbone pre-trained weights frozen in their experiments. 
However, we train end-to-end to leverage the information obtained from the captions to influence the Swin backbone weights.

\noindent \textbf{Object Detection and Transformer-based vision backbones:} 
Object detection research has seen a breakthrough with the introduction of CNN-based models like R-CNN~\cite{RCNN}, Fast R-CNN~\cite{Fast-RCNN}, and Faster R-CNN~\cite{Faster-RCNN}. Later, the remarkable success of the Transformer architecture~\cite{Attention_is_all_you_need} in the NLP domain has motivated researchers to explore its application in computer vision. Transformers were first introduced for vision problems in Vision Transformers (ViT)~\cite{ViT}. Several works on ViT and its variants followed up~\cite{Deit,TiT,pyramid_vision_transformer}. 
However, ViT required large-scale training datasets like JFT-300M to perform optimally. DeiT~\cite{Deit} addresses this limitation by introducing training strategies that enable ViT to work effectively with smaller datasets like ImageNet-1K. While ViT shows promising results in image classification, it is not suitable as a general-purpose backbone network for dense vision tasks or high-resolution input images~\cite{Swin_transformer}. 
In parallel to Swin Transformer~\cite{Swin_transformer}, other researchers have also modified the ViT architecture to improve image classification~\cite{TiT}. But Swin Transformer \cite{Swin_transformer} demonstrates that they achieve the best speed-accuracy trade-off among the above-mentioned methods on image classification, though it focuses on a general-purpose backbone. Also, Swin Transformer has linear complexity to image size unlike \cite{pyramid_vision_transformer} with quadratic complexity.  

\noindent \textbf{Multitask Learning:}
Early works on multitask image captioning such as  \cite{multitask_img_caption_Neurips_16} incorporate a simple multi-label  classification as an auxiliary task for image captioning. \cite{multitask_img_caption_IJCAI_18}~expands on this by introducing two auxiliary tasks: multi-label classification and syntax generation. Both papers highlight the benefits of auxiliary tasks in enhancing the performance of image captioning and use CNN-LSTM as an encoder-decoder. In the proposed approach, we use a more complex object detection as the auxiliary task and use Swin Transformer in our architecture. Recently, there has been a growing interest in developing models which can solve multiple tasks together. For instance, Pix2seq-v2~\cite{chen22} proposes a prompt-based approach in an encoder-decoder framework based on Transformers to solve four tasks, namely, object detection, instance segmentation, keypoint detection, and image captioning.
\begin{figure*}[ht]
    \centering
    \vspace{-4mm}
    \includegraphics[width=0.98\textwidth]{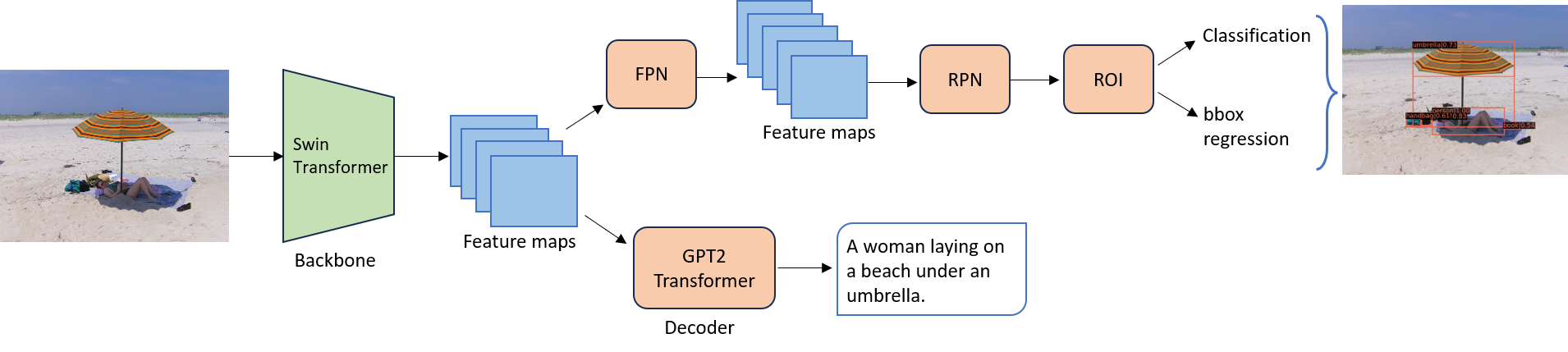}
    \vspace{-2mm}
    \caption{\small Architectural overview of the proposed \textbf{T}ransformer-based \textbf{I}mage \textbf{C}aptioning and \textbf{O}bject \textbf{D}etection  (\textbf{TICOD}) model.}
    \label{fig:archi}
    \vspace{-10mm}
\end{figure*}

\section{Proposed Method}
\label{sec_Method}
\vspace{-2mm}
This work aims to leverage multitask learning to train objection detection and image captioning tasks jointly.
The overall architecture of our model is shown in Fig~\ref{fig:archi}.  The complete architecture has three major components: (a) an initial image feature extractor, (b) an object identification network, and (c) a caption generation network. As the name suggests, the image feature extractor extracts features from the input image. These features are then passed to the two networks for object detection and caption generation. Swin Transformer~\cite{Swin_transformer} is used for the image feature extractor owing to its superior performance in various image understanding tasks \cite{Swin_transformer,swinIR,PureT}.


The object detection network involves passing the extracted image features through a sequence of Feature Pyramid Network (FPN) \cite{FPN}, Region Proposal Network (RPN) \cite{Faster-RCNN}, and Region of Interest (RoI) pooling layer \cite{Fast-RCNN} to get the objects and their bounding boxes detected. It can be seen that this flow of encoding the image representation to finally detecting the objects follows the overall framework of Faster R-CNN \cite{Faster-RCNN}. This specific instantiation of Faster R-CNN uses Swin Transformer as the backbone to extract the image feature maps. On the other hand, the caption generation network passes the extracted image features through a GPT-2 \cite{GPT2} architecture to get the final captions. These two networks operate in parallel to solve the tasks of object detection and caption generation and the losses for the tasks are combined into a multitask loss. The use of a common feature extractor for both the tasks, and a combined loss function to guide the training process enables the two tasks to influence the learning of each other and improve their individual performance.

As shown in Fig.~\ref{fig:archi},  our model consists of a common encoder (i.e., image feature extractor) and two parallel decoders (i.e., object identification network and caption generation network),  trained as a single model with a joint loss function. Given an input image, it is first divided into image patches with patch size $= 4 \times 4.$ It is then passed through a Swin backbone, producing four feature maps of the following dimensions:

\begin{equation*}
\scriptsize
    \begin{split}
        \biggl\{ \Bigl( bs \times C \times \frac{H}{4} \times \frac{W}{4} \Bigl),  
        \Bigl( bs \times 2C \times \frac{H}{8} \times \frac{W}{8} \Bigl), 
         \Bigl( bs \times 4C \times \frac{H}{16} \times \frac{W}{16} \Bigl), 
        \Bigl( bs \times 8C \times \frac{H}{32} \times \frac{W}{32} \Bigl) \biggl\}
    \end{split}
\end{equation*}
where,  $bs$ denotes the batch size, $C$ denotes the channel dimensions, i.e., the patch embedding dimension in each Swin block, and $H, W$ denote the height and width of the image, respectively. We call this as the \textit{backbone network}.

Each of these feature maps is passed through the FPN regarded as the \textit{neck} in object detection literature. It produces five feature maps, each of embedding dimension$=256$~\cite{FPN}. FPN is used at multiple levels to increase the detection of small objects. It provides semantically rich information at multiple scales, which reduces the error rate in detection \cite{FPN}. Feature maps from the FPN are passed into the RPN \cite{Faster-RCNN} and RoI pooling layer \cite{Fast-RCNN,Faster-RCNN} to finally produce the object classes and bounding box coordinates. We refer to this as the \textit{object detection network}.

The other parallel branch of our multitask model, which we call the \textit{caption network}, consists of a GPT2 \cite{GPT2} Transformer as the decoder. It takes the last feature map of the Swin backbone as input along with a \texttt{<start>} token and generates captions word by word in an auto-regressive manner.

\noindent\textbf{Attention:} We use Multi-head Self Attention (MSA)~\cite{Attention_is_all_you_need} in the decoder to calculate the relationship between tokens in a sequence and cross-attention for the relationship between the tokens and image grid features. We also adopt Window MSA/Shifted Window MSA (W-MSA/SW-MSA) proposed in the Swin Transformer work~\cite{Swin_transformer}. They are used in the encoder to model the relationship between the image patches.
\vspace{-3mm}
\subsection{Objective function}
\label{obj_func}
For training the caption network, we use the standard language modeling loss, which is the Cross-Entropy loss ($L^C(\theta,\phi)$) computed over all the samples. For a given sample, the loss function aims to predict the next token given the context and can be mathematically formulated as follows:
\begin{equation} \label{cross-entropy_loss}
    L^C (\theta, \phi) = -\sum_{t=1}^{T} log(p(y_t ^* \mid y_{1:t-1} ^*,x))
\end{equation}
where $\theta$ and $\phi$ represent the parameters of Swin Transformer backbone and GPT2 respectively, $y_{1:T} ^*$ is the target ground-truth sequence, and $x$ is the input image. 

For objection detection, we consider a loss ($L^O(\theta,\psi)$) which consists of the Cross-Entropy Loss for object classification and smooth $L1$ Loss \cite{Fast-RCNN} for bounding box regression computed over all the samples. Here. $\psi$ represents the parameters of the Faster R-CNN/Cascade R-CNN network used for object detection.  Since the \textit{object detection network} consists of the Region Proposal Network (RPN) and Region of Interest (RoI) pooling layer, $L^O$ can be further subdivided  as -
\begin{equation}
\begin{aligned}
    L^O(\theta,\psi) 
       &= L_{cls} ^{RPN} + L_{reg} ^{RPN} + L_{cls} ^{RoI} + L_{reg} ^{RoI} 
\end{aligned}
\end{equation}
where $L^{RPN}$ and $L^{RoI}$ denote the losses from RPN and RoI layers respectively. 
$L_{cls} ^{RPN}$ and $L_{reg} ^{RPN}$ denote classification and bounding box (bbox) regression losses from the RPN network. $L_{cls} ^{RoI}$ and $L_{reg} ^{RoI}$ refer to the classification loss and bbox regression loss from the RoI network. $L_{cls} ^{RPN}$ is a classification log loss over two classes -- object or non-object, i.e., \textit{background} class \cite{Faster-RCNN}. $L_{reg} ^{RPN}$ is a Smooth $L_1$ Loss defined in \cite{Fast-RCNN} arising from the difference between the ground-truth bbox coordinates and the predicted bbox coordinates \cite{Faster-RCNN}. 
$L_{cls} ^{RoI}$ is the log loss over $(C+1)$ categories -- $C$ object classes and a \textit{background} class, and $L_{reg} ^{RoI}$ is again a similar Smooth $L_1$ Loss over the bbox coordinates~\cite{Fast-RCNN}.

For training our model. we use the joint multitask learning objective function, which combine the image captioning and object detection losses as 
\begin{equation}
    L^T (\theta, \phi,\psi) = L^O(\theta, \psi) + \lambda \cdot L^C(\theta, \phi)
    \label{eq: loss_func}
\end{equation}
where $\lambda$ is the weightage to be given to the captioning loss.
The $\lambda$ value is  chosen to maximize the evaluation scores of both image captioning and object identification. It is determined empirically using a validation data as demonstrated in Table~\ref{table: choosing_lambda}. We found the most suitable values of $\lambda$ to be  $0.1$ and $0.2$ for TICOD-small and TICOD-large, respectively.
\vspace{-4mm}
\section{Experimental Setup}
\label{sec_experiments}
\vspace{-2mm}
\noindent \textbf{Dataset:}
\noindent We use the MS-COCO 2017 dataset \cite{COCO_caps,COCO_objects} containing 118K training and 5K validation images. COCO has five captions per image and separate annotations for object categories and bounding boxes. For comparison with image captioning works, we follow the standard ``Karpathy'' split.

\noindent \textbf{Evaluation Metrics:}
\noindent \textit{Image Captioning--} We evaluate our captioning performance using the standard metrics used in Natural Language Processing, viz, BLEU, CIDEr, METEOR, ROUGE-L, and SPICE. The metrics above only evaluate the generated captions by matching them with the reference captions at the lexical level. Since these metrics try to find exact matches, the scores might not be a true measure of the quality of the captions, as the presence of synonymous words may lower the scores. So, we also measure the scores based on BERTScore \cite{bert-score}, which uses contextual embeddings to find a semantic similarity measure between the candidate sentence and the reference sentence. BERTScore has been shown to exhibit a superior correlation with human judgments \cite{support_bertscore}, and provide strong model selection performance \cite{bert-score}.

\noindent \textit{Object Detection--}  We use the standard evaluation metrics -- mAP as the mean of APs@$[.5:.05:.95]$, AP@$IoU=0.50$, and AP@$IoU=0.75$. We also report the APs across scales, i.e., APs@small, medium, and large objects.

\noindent \textbf{Implementation Details}: We keep the same settings as in Swin Transformer \cite{Swin_transformer} work. 
AdamW optimizer is used with an initial learning rate of $10^{-4}$, weight decay of 0.05, and batch size of 2 per GPU. 
For training, we use 4 Nvidia V100 GPUs and for inference, we use a single V100 GPU. During inference, captions are generated using beam search with beam size $= 5$.
For the image captioning part, we take the encoding from the last feature map of the Swin backbone. The last feature map of Swin-T and Swin-B models have embedding dimensions of $8 \times 96 = 768$ and $8 \times 128 = 1024$, respectively~\cite{Swin_transformer}, which matches the embedding dimensions of GPT2-small $(dim=768)$ and GPT2-medium $(dim=1024)$. The combined loss from the object detection and the caption networks is backpropagated to update the model weights of the backbone network.

\noindent \textbf{Architecture Details}:
The authors of Swin Transformer \cite{Swin_transformer} have proposed several variants of the Swin backbone: Swin - tiny, small, base, and large. For our experiments, we have used --
\begin{itemize}
    \item Swin-tiny \cite{Swin_transformer} with GPT2-small \cite{GPT2} for image captioning and Swin-tiny backbone in Faster R-CNN framework~\cite{Faster-RCNN} for object detection. We call this \textit{TICOD-small}.
    \item Swin-base \cite{Swin_transformer} with GPT2-medium \cite{GPT2} for image captioning and Swin-base backbone in Cascade R-CNN  \cite{cascade_rcnn} for object detection. We call this as \textit{TICOD-large}.
\end{itemize}
Our model TICOD-large performs better than the other variant. The Swin Transformer acts as a common backbone for image captioning and object detection in our multitask model. 
\vspace{-4mm}
\section{Results}
\label{sec_results}
\vspace{-2mm}
\subsection{Comparison and Analysis}
\label{sec_comparision}
\vspace{-2mm}
We compare our multitask model with task-specific baselines from the literature on image captioning and object detection. For image captioning, we compare our work with some recent models like -- ClipCap~\cite{clipcap}, Meshed-memory ($\mathcal{M}^2$) Transformer~\cite{M2_transformer}, and PureT~\cite{PureT} on the MS-COCO~\cite{COCO_caps} offline test split.  We also compare our work with  Pix2seq-V2~\cite{chen22}, a recently proposed multitask system that jointly learns object detection and image captioning along with two other vision tasks. 
Table~\ref{table: comparision_SOTA} reports the performances of these models and our proposed model. 
\begin{table}[ht]
\vspace{-5mm}
   \input{tables/table1_comparison_SOTA}
   \vspace{-2mm}
  \caption{\small Comparison on MS-COCO \cite{COCO_caps,COCO_objects} dataset.}
   \label{table: comparision_SOTA}
\vspace{-9mm}
\end{table}
From Table~\ref{table: comparision_SOTA}, it can be seen that the PureT~\cite{PureT} model outperforms all other models in terms of the BLEU (B1 to B4), Meteor, RougeL, CIDEr, and Spice metrics. However, these are all lexical similarity based metrics, and try to match the exact words present in the generated and ground-truth captions. Although a good score in terms of these metrics indicate good match between the generated and the reference captions, a lower score does not necessarily indicate a poor match. This is because any concept or thought can be expressed in multiple ways with very less overlap in the words used in these \textit{parallel expressions}. This drawback is addressed in the BERTScore metric~\cite{bert-score}, where the semantic embeddings of the texts are compared to decide the performance score. Our model achieves comparable performances with PureT and $\mathcal{M}^2$ transformer in terms of the lexical overlap based scores. At the same time, it achieves $\textbf{3.65\%}$ and $\textbf{9.66\%}$ improvements in BERTScore over these two models respectively, indicating that the proposed model can generate better quality captions at a semantic level. The proposed TICOD-large model also outperforms both model variants of ClipCap \cite{clipcap} in terms of all the metrics as well as BERTScore. The mentioned BERTScores in this table are calculated using Deberta-xlarge-mnli model~\cite{deberta} as it best correlates with human evaluation~\cite{bert-score}. Our model also outperforms the popular BUTD model~\cite{Anderson_Bottom-up} in terms of CIDEr, Meteor, RougeL, Spice. 

In addition, we also calculate BERTScore with other models like -- Roberta-large~\cite{roberta} and Deberta-xlarge-mnli with idf (Inverse Document Frequency). BERTScore calculated using these three models for the baselines and the proposed multitask model are illustrated in Table~\ref{table: BERTScore}. Clearly, our proposed multitask model outperforms all the baselines in terms of BERTScore calculated using all three mentioned methods. Comparable scores based on lexical overlap-based metrics and superior scores based on embedding-based methods indicate that the proposed model can generate good-quality captions for the input images.
\begin{table}[ht]
\vspace{-6mm}
   \input{tables/BERTScore_table}
   \vspace{-2mm}
  \caption{\small Comparison of captioning performance of our proposed multitask model with some image captioning baselines from literature, in terms of BERTScore \cite{bert-score}.}
   \label{table: BERTScore}
   \vspace{-10mm}
\end{table}

We report the object detection evaluation scores also in Table~\ref{table: comparision_SOTA} for convenience of comparison. Our objective of this work is to perform image captioning and object detection simultaneously by improving the performance of image captioning due to joint training with a carefully constructed joint loss function. We demonstrate that we achieve superior image captioning performance in terms of BERTScore while maintaining a comparable performance in object detection. Since our model is developed upon Swin Transformer architecture \cite{Swin_transformer}, we compare our model's performance on object detection with Swin Transformer \cite{Swin_transformer} in the Cascade R-CNN \cite{cascade_rcnn} framework. The comparisons presented in Table~\ref{table: comparision_SOTA}, show that TICOD has better performance in terms of mAP, AP@0.75, AP@small, and AP@medium objects. This, in turn, shows that the caption generation task has positively influenced the object detection task through the joint training, and has resulted in improved performance for object detection. We also compare our model with other popular baselines like DETR~\cite{DETR} and Faster R-CNN~\cite{Faster-RCNN}. 
We compare with Pix2seq-v2~\cite{chen22} on their reported object detection and image captioning scores, and we can see a clear improvement in performance using our approach for both tasks. This is because Pix2seq-v2 doesn't use detection-specific architecture but instead uses language modeling to solve ``core" vision tasks. Due to the significant departure from conventional architectures, the model needs further improvement to challenge the current SOTA of task-specific models~\cite{chen22}. It also has a slower inference speed (particularly for longer sequences) than the specialized systems, as the approach is based on autoregressive modeling~\cite{chen22}. 

\begin{figure}[!]
\centering
\vspace{-6mm}
\begin{subfigure}[c]{.24\textwidth}
\includegraphics[width=0.675\linewidth,angle=270,origin=c]{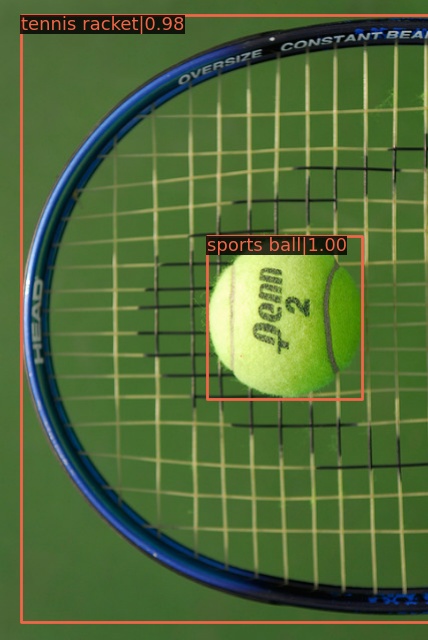}
  \vspace{-9mm}
  \caption{}
  { \begin{footnotesize}
  \textbf{GT1:} A tennis ball sitting on top of a tennis racket.\\
  \textbf{GT2:}  A tennis ball is sitting on a tennis racket.\\
    \textbf{$\mathcal{M}^2:$}A tennis ball on a green tennis racket on a\\
    \textbf{PureT:} A tennis racket and a tennis ball on a table.\\
    \textbf{Ours:} A tennis ball sitting on top of a racquet.
    \end{footnotesize}}
  \label{fig:img1}
\end{subfigure}
\hfill
\begin{subfigure}[c]{.24\textwidth}
  \includegraphics[width=1.0\linewidth]{images/COCO_img_obj/000000193884.jpg} 
  \caption{}
  {\begin{footnotesize}
  \textbf{GT1:} A red stop sign sitting on the side of a road.\\
  \textbf{GT2:} Stop sign on a street of a cemetary.\\
    \textbf{$\mathcal{M}^2:$} A stop sign on the side of a street.\\
    \textbf{PureT:} A stop sign on the side of a street.\\
    \textbf{Ours:} A \textit{red} stop sign sitting on the side of a road.
    \end{footnotesize}}
  \label{fig:img2}
\end{subfigure}
\hfill
\begin{subfigure}[c]{.24\textwidth}
  \includegraphics[width=1.0\linewidth] {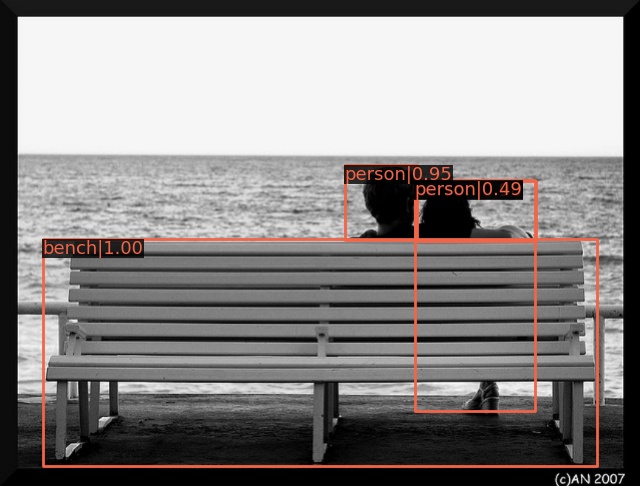}
  \caption{}
  {\begin{footnotesize}
  \textbf{GT1:} Two people are sitting on a bench together in front of water.\\
  \textbf{GT2:} A couple is sitting on a bench in front of the water.\\
    \textbf{$\mathcal{M}^2:$} Two people sitting on a bench near the water.\\
    \textbf{PureT:} Two people sitting on a bench looking at the ocean.\\
    \textbf{Ours:} Two people sitting on a bench facing the ocean.
     \end{footnotesize}
    }
  \label{fig:img3}
\end{subfigure}
\hfill
\begin{subfigure}[c]{.24\textwidth}
  \includegraphics[width=1.0\linewidth]{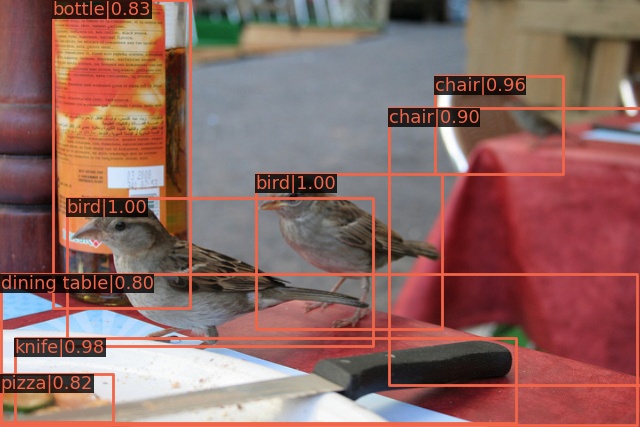} 
  \caption{}
  {
  \begin{footnotesize}
  \textbf{GT1:} A couple of small birds standing on top of a table.\\
  \textbf{GT2:} Two little sparrows standing on a table by a knife.\\
    \textbf{$\mathcal{M}^2:$} A small bird sitting on a table next to a knife.\\
    \textbf{PureT:} Two birds sitting on top of a plate of food.\\
    \textbf{Ours:} A couple of small birds standing on top of a table.
    \end{footnotesize}
    }
  \label{fig:img4}
\end{subfigure}
  \caption{\small Examples of captions generated by  $\mathcal{M}^2$-Transformer~\cite{M2_transformer}, PureT~\cite{PureT}, our model, and their corresponding ground-truths(GT)~\cite{COCO_caps}. The images also display the detected object categories and their scores as predicted by our proposed model.}
  \label{fig:qualitative_example_images}
  \vspace{-7mm}
\end{figure}

\noindent \textbf{Qualitative Comparison:}
While discussing the caption generation performance of the different models, based on the values of the evaluation metrics, we argued that TICOD generates good-quality captions. However, it may use words that are semantically similar but lexically different from the tokens in the corresponding ground-truth captions. Fig.~\ref{fig:qualitative_example_images} presents some example cases to elaborate on this point. It shows a few images with detected objects by our proposed model and their corresponding captions, along with the captions generated by PureT model~\cite{PureT}, $\mathcal{M}^2$ Transformer~\cite{M2_transformer} and the ground-truths. In Fig.~\ref{fig:img1}, our model generates the correct caption whereas $\mathcal{M}^2$ incorrectly generates \textit{green} tennis racket. The generation also doesn't end properly as it produces \textit{on a} at the end. PureT generated \textit{on a table}, which is incorrect. In Fig.~\ref{fig:img2}, our model generates \textit{red stop sign} which is more detailed than $\mathcal{M}^2$ and PureT model's captions. In Fig.~\ref{fig:img3}, the captions are all similar. However, in Fig.~\ref{fig:img4}, we notice that our model has slightly under-performed as it generates \textit{A couple of small birds} instead of \textit{Two birds} produced by PureT. $\mathcal{M}^2$ also under-performed by producing \textit{A small bird}. By qualitatively analyzing the captions produced by our model, we have observed that, while in most cases our model produces better or equivalent captions, there are some cases where our model has slightly regressed performance.
\vspace{-4mm}
\subsection{Ablation Studies}
\label{ablation}
\vspace{-2mm}
To evaluate the effectiveness of our proposed approach, we perform ablation studies on the COCO dataset \cite{COCO_caps,COCO_objects} using two object detection frameworks -- Faster R-CNN \cite{Faster-RCNN} and Cascade R-CNN \cite{cascade_rcnn}. We also check the model's performance by keeping different parts of the network frozen.

\begin{table}[ht]
\vspace{-8mm}
   \input{tables/ablation}
   \vspace{-2mm}
  \caption{\small Ablation study with different methods, backbones and decoder sizes on COCO \cite{COCO_caps,COCO_objects} dataset.}
   \label{table:ablation}
   \vspace{-11mm}
\end{table}
\noindent \textbf{Effect of backbone size and object detection framework: }
Table~\ref{table:ablation} shows the performance of our model with different backbone sizes and object detection frameworks. As observed, our model performs better when Cascade R-CNN is used with Swin-base and GPT2-medium. There is a performance gain of $\sim \textbf{3\%}$ Bleu-1 score, $\textbf{21.67\%}$ Bleu-4 and $\textbf{21.09\%}$ CIDEr ($\textbf{+18.5}$ CIDEr) over Faster-RCNN with Swin-tiny and GPT2-small with frozen backbone and object detection network (fine-tuning only the decoder network). Also, there is an improvement of $\textbf{6.63\%}$ Bleu-1, $\textbf{27\%}$ Bleu-4, and $\textbf{28\%}$ CIDEr ($ \textbf{+ 25.2}$ CIDEr) when the backbone and decoder networks are finetuned. Clearly, with larger backbone and decoder sizes, there is performance improvement.

\noindent \textbf{Effect of frozen layers:} We perform experiments by keeping different components of the network frozen. The trainable components of the network are mentioned in the fourth column of Table \ref{table:ablation}. The remaining components of the model were frozen. We tried three combinations of $\langle$frozen-finetuned$\rangle$ components: (i) Swin backbone, FPN, RPN, ROI layers frozen while finetuning only the GPT2 decoder, (ii) finetuning Swin backbone and GPT2 while keeping the remaining layers required for object detection as frozen, and (iii) finetuning all components by keeping the caption network turned off. It is observed from Table~\ref{table:ablation} that case (ii) has a performance improvement of  $\textbf{2.74\%}$ CIDEr for Faster R-CNN method and $\textbf{8.57\%}$ CIDEr for Cascade R-CNN. BERTScore has also improved by $\textbf{3.65\%}$ for Cascade R-CNN. We also observe that for object detection, the mAP improves from 51.9 to $\textbf{52.1}$ when the backbone network is trainable, which demonstrates the positive impact of joint training.

For both the upper and lower halves of Table \ref{table:ablation}, the last rows represent the setup where the GPT-2 network is frozen, and the object detection network is finetuned. In the proposed TICOD model, the caption network is passed an image embedding. However, the GPT-2 in the caption network is initialized with pre-trained GPT-2 parameters that do not recognize that image embedding as input. Hence, without finetuning the caption-generation network, the generated captions for this setup would be meaningless. Accordingly, the values of the evaluation metrics for image captioning are filled by ``--"s  for these rows.

The object detection scores in the first and third rows are identical in both halves of the table. The first row corresponds to a setup where the object detection network is frozen, while the third row involves finetuning the parameters of the object detection network. The similarity in scores arises from the fact that the object detection network's parameters are initialized with a pre-trained checkpoint optimized for this dataset. Further finetuning leads to a drop in performance on the validation set. As a result, the initial parameters (epoch 0) are retained, resulting in similar metric values for these rows for the object detection task.

 \noindent \textbf{Task-specific performance: }
 We examine the impact of multitasking on performance compared to task-specific models through experiments with two baselines: (i) object detection alone without the caption network, and (ii) captioning alone without the object detection network. Results in Table~\ref{table:ablation} indicate improved image captioning performance without compromising object detection.
 
\noindent \textbf{Choosing Lambda: } 
We conduct experiments to finetune the hyperparameter $\lambda$ in Equation~\ref{eq: loss_func}. Table~\ref{table: choosing_lambda} demonstrates that $0.1$ and $0.2$ are the most suitable values for TICOD-small and TICOD-large, respectively. For TICOD-small, the difference in image captioning scores between $\lambda = 0.1$ and $10$ is negligible, and as object detection performance degrades with increasing lambda, scores for $\lambda = 0.2$ and $0.5$ are not reported.
\begin{table}[ht]
 \vspace{-7mm}
   \input{tables/choosing_lambda}
   \vspace{-2mm}
  \caption{\small \textbf{Hyperparameter tuning:} Illustration of model performance with different $\lambda$ values.}
   \label{table: choosing_lambda}
   \vspace{-8mm}
\end{table}
\vspace{-6mm}
\section{Conclusion}
\label{sec_conclusion}
\vspace{-2mm}
In this work, we presented TICOD, a multitask framework for object detection and image captioning. Empirically, we show that joint learning helps improve image captioning by improving the image representations in the backbone. Swin Transformer is not pre-trained on a vision-language objective, yet we demonstrate that we can use it directly with GPT2 and show superior image captioning performance in BERTScore while maintaining a comparable performance in object detection. Our proposed framework is customizable as Swin Transformer and GPT2 can be replaced with newer specialized SOTA detection and large language models, which will further improve performance over general-purpose multitask models like Pix2seq-V2~\cite{chen22}. 
 
\noindent \textbf{Acknowledgements:} This work was supported by DST National Mission on Interdisciplinary Cyber-Physical Systems (NM-ICPS), Technology Innovation Hub on Autonomous Navigation and Data Acquisition Systems: TiHAN Foundations at Indian Institute of Technology (IIT) Hyderabad, India. We also acknowledge the support from Japan International Cooperation Agency (JICA). We express gratitude to Suvodip Dey for his valuable insights and reviews on this work.
\vspace{-5mm}
\bibliographystyle{splncs04}
 \bibliography{mybib}

\end{document}

%% file: tables/table1_comparison_SOTA.tex
\begin{center}
    \resizebox{\textwidth}{!}{%
    \begin{tabular}{l|cccccccc|c|cccccc} 
    \toprule
    Methods &B1 &B2 &B3 &B4 &Meteor &RougeL &CIDEr &Spice &BERTScore &mAP &AP$_{50}$ &AP$_{75}$ &AP$_S$ &AP$_M$ &AP$_L$ \\ \midrule
    ClipCap (MLP+GPT2 fine-tuning) \cite{clipcap} 
    &70.9 &54.4 &41.2 &31.5 &27.7 &54.8 &106.7 &20.5 &68.001 &-- &-- &-- &-- &-- &-- \\
    ClipCap (Transformer)~\cite{clipcap} 
    &74.6 &58.5 &44.5 &33.5 &27.6 &55.9 &112.8 &21.0 &68.326 &--&--&--&--&--&--\\ 
    $\mathcal{M}^2$- Transformer \cite{M2_transformer} &80.8	&--	&-- &39.1 &29.2 &58.6 &131.2 &22.6 &64.556 &-- &-- &-- &-- &-- &--\\
    PureT \cite{PureT} &\textbf{82.1}
    &\textbf{67.3} &\textbf{53.0} &\textbf{40.9} &\textbf{30.2} &\textbf{60.1} &\textbf{138.1} &\textbf{24.2} &68.303 &-- &-- &-- &-- &-- &-- \\
    BUTD \cite{Anderson_Bottom-up} &77.2 &-- &-- &36.2 &27.0 &56.4 &113.5 &20.3&-- &-- &-- &-- &-- &-- &-- \\
    Pix2Seq-V2 \cite{chen22} &-- &-- &-- &34.9 &-- &--&-- &-- &-- &46.5 &-- &-- &-- &-- &--\\
    \midrule
    Swin-B (Cascade R-CNN) \cite{Swin_transformer} &--&--&--&--&--&--&--&--&-- & 51.9 &\textbf{70.9} &56.5 &35.4 &55.2 & \textbf{67.3}\\
    DETR-R101 \cite{DETR} 
    &--&--&--&--&--&--&--&--&-- &43.5 &63.8 &46.4 &21.9 &48.0 &61.8\\
    Faster R-CNN R101-FPN \cite{Faster-RCNN}  &--&--&--&--&--&--&--&--&--&42.0 &62.5 &45.9 &25.2 &45.6 &54.6\\
    \midrule
    TICOD-large (Ours) 
    &75.6 &59.0 &45.5 &35.3 &28.3 &56.7 &115.3 &21.1 &\textbf{70.794} &\textbf{52.1} &70.6 &\textbf{56.7} &\textbf{34.8} &\textbf{55.3} & 67.2\\
    \bottomrule
    \end{tabular}%
    }
\end{center}

%% file: tables/BERTScore_table.tex
\begin{center}
\resizebox{ 0.65\textwidth}{!}{
    \begin{tabular}{m{5.7cm}|m{1.2cm}|m{1.2cm}|m{1.5cm}} 
    \toprule
     Methods &Deberta-xlarge-mnli \cite{deberta} &Roberta-large \cite{roberta} &Deberta-xlarge-mnli with idf \\
     \midrule
     $\mathcal{M}^2$- Transformer \cite{M2_transformer} &  64.56 &63.11 &59.40\\
    ClipCap (MLP+GPT2 fine-tuning) \cite{clipcap} &68.00 &65.27 &59.13\\
    ClipCap (Transformer)~\cite{clipcap} &68.33 &66.25 &59.37\\
    PureT \cite{PureT} &68.30 &67.69 &63.22\\
    \midrule
    TICOD-large ($\lambda$ = 0.2) (proposed model) &70.79 &68.06 &63.23\\
    TICOD-large ($\lambda$ = 0.5) &71.69 &68.98 &63.40\\
    \bottomrule
    \end{tabular} 
    }
\end{center}

%% file: tables/ablation.tex
\begin{center}
    \resizebox{\textwidth}{!}{%
    \begin{tabular}{p{1.2cm}|c|c|p{1.4cm}|cccccccc|c|cccccc}
        \toprule
         Methods &Backbone &Decoder & Finetuned networks &B1 &B2 &B3 &B4 &Meteor &RougeL &CIDEr &Spice &BERTScore &mAP &AP$_{50}$ &AP$_{75}$ &AP$_S$ &AP$_M$ &AP$_L$ \\ 
         \midrule
         \multirow{3}{1.2cm}{Faster R-CNN \cite{Faster-RCNN}}  & \multirow{3}{*}{Swin-T} & GPT2 (small) &GPT2 &70.4 &51.5 &36.8 &26.3 &23.7 &50.9 &87.7 &16.7 &66.473 &\textbf{46.0} &\textbf{68.1} &\textbf{50.3} &\textbf{31.2} &\textbf{49.2} &\textbf{60.1}\\
         & & GPT2 (small) & Swin,GPT2  
         &\textbf{70.9} &\textbf{52.5} &\textbf{38.2} &\textbf{27.8} &\textbf{24.1} &\textbf{51.3} &\textbf{90.1} &\textbf{17.1} &\textbf{66.934} &45.3 &67.7 &49.9 &29.5 &48.7 &59.0\\
         & & \xmark & Swin,FPN, RPN, ROI  
         &-- &-- &-- &-- &-- &-- &-- &-- &-- &\textbf{46.0} &\textbf{68.1} &\textbf{50.3} &\textbf{31.2} &\textbf{49.2} &\textbf{60.1}\\
         \midrule
         \multirow{3}{1.2cm}{Cascade R-CNN \cite{cascade_rcnn}} & \multirow{3}{*}{Swin-B} &GPT2 (med) & GPT2  
         &72.5 &55.3 &41.8 &32.0 &27.2 &54.6 &106.2 &19.7 &68.301 &51.9 &\textbf{70.9} &56.5 &\textbf{35.4} &55.2 &\textbf{67.3}\\
         & & GPT2 (med) & Swin,GPT2 
         &\textbf{75.6} &\textbf{59.0} &\textbf{45.5} &\textbf{35.3} &\textbf{28.3} &\textbf{56.7} &\textbf{115.3} &\textbf{21.1} &\textbf{70.794} &\textbf{52.1} &70.6 &\textbf{56.7} &34.8 &\textbf{55.3} &67.2\\
         & & \xmark & Swin,FPN, RPN, ROI  
         &-- &-- &-- &-- &-- &-- &-- &-- &-- &51.9 &\textbf{70.9} &56.5 &\textbf{35.4} &55.2 &\textbf{67.3}\\ 
         \bottomrule
    \end{tabular}
    }
\end{center}

%% file: tables/choosing_lambda.tex
\begin{center}
    \resizebox{\textwidth}{!}
    {
    \begin{tabular}{l|c|cccccccc|c|cccccc}
        \toprule
         Methods &lambda ($\lambda$) &B1 &B2 &B3 &B4 &Meteor &RougeL &CIDEr &Spice &BERTScore &mAP &AP$_{50}$ &AP$_{75}$ &AP$_S$ &AP$_M$ &AP$_L$ \\ 
         \midrule
         \multirow{3}{*}{TICOD-small} & 0.01 &69.7 &51.3 &36.6 &25.8 &22.8 &50.1 &82.1 &16.2 &65.728 &\textbf{45.4} &67.3 &\textbf{50.2} &\textbf{31.1} &\textbf{48.8} &\textbf{59.1}\\
         & 0.1
         &\textbf{70.9} &\textbf{52.5} &\textbf{38.2} &\textbf{27.8} &\textbf{24.1} &\textbf{51.3} &\textbf{90.1} &\textbf{17.1} &\textbf{66.934} &45.3 &\textbf{67.7} &49.9 &29.5 &48.7 &59.0\\
         &10 
         &70.8 &52.8 &38.1 &27.4 &23.8 &51.3 &88.6 &17.2 &66.698 &37.7 &62.3 &40.6 &25.8 &42.0 &46.1\\
         \midrule 
         \multirow{5}{*}{TICOD-large} & 0.01 &72.8 &56.9 &41.8 &32.1 & 26.7 &54.2 & 100.7 &18.6 &67.801 &51.9 &70.5 &56.6 &\textbf{36.1} &55.2 & \textbf{67.4}\\
         &0.1 &74.3 &57.3 &43.6 &33.5 &27.0 &55.1 &107.3 &19.4 &69.821 &51.9 &\textbf{70.6} &56.5 &\textbf{36.1} &\textbf{55.3} &67.2\\
         &0.2 &75.6 &59.0 &45.5 &35.3 &28.3 &56.7 &115.3 &21.1 &70.794 &\textbf{52.1} &\textbf{70.6} &\textbf{56.7} &34.8 &\textbf{55.3} &67.2\\
         &0.5 &\textbf{76.5} &60.4 &\textbf{46.9} &\textbf{36.6} &\textbf{28.9} &\textbf{57.6} &\textbf{119.6} &\textbf{21.6} &\textbf{71.686} &51.6 &70.3 &56.3 &33.6 &55.0 &\textbf{67.4}\\

         &10 &76.2 &\textbf{60.5} &45.7 &36.1 &28.5 &57.1 &119.4 &21.5 &71.221 &47.2 &67.8 &50.4 &30.2 &51.5 &61.1\\
         \bottomrule
    \end{tabular}
}
\end{center}

%% file: paper-491.bbl
\begin{thebibliography}{10}
\bibitem{GPT2}
Radford, A., et~al.: Language models are unsupervised multitask learners. OpenAI blog  ('19)
\end{thebibliography}
